\documentclass[11pt]{article}

\usepackage[]{EMNLP2022}

\usepackage{times}
\usepackage{latexsym}
\usepackage{marvosym}
\usepackage[T1]{fontenc}

\usepackage[utf8]{inputenc}

\usepackage{microtype}

\usepackage{inconsolata}

\usepackage{graphicx}

\usepackage{CJKutf8} 

\usepackage{amsmath} 

\usepackage{booktabs}

\usepackage{multirow}

\usepackage{makecell}

\title{Title2Event: Benchmarking Open Event Extraction with \\a Large-scale Chinese Title Dataset}

\author{Haolin Deng\textsuperscript{$1$} \enskip Yanan Zhang\textsuperscript{$1$}\thanks{\;\;Corresponding author} \enskip Yangfan Zhang\textsuperscript{$1$} \enskip Wangyang Ying\textsuperscript{$1$} \\
\textbf{Changlong Yu}\textsuperscript{$2$} \enskip \textbf{Jun Gao} \enskip \textbf{Wei Wang}\textsuperscript{$3$} \enskip \textbf{Xiaoling Bai}\textsuperscript{$1$} \\
\textbf{Nan Yang}\textsuperscript{$1$} \enskip \textbf{Jin Ma}\textsuperscript{$4$} \enskip \textbf{Xiang Chen}\textsuperscript{$1$} \enskip \textbf{Tianhua Zhou}\textsuperscript{$1$} \\
 \textsuperscript{$1$}Tencent \quad \textsuperscript{$2$}HKUST \quad \textsuperscript{$3$}Tsinghua University  \quad \textsuperscript{$4$}USTC\\
   hldeng028@gmail.com, \ \{yananzhang, devinbai\}@tencent.com \\
}

\begin{document}
\begin{CJK*}{UTF8}{gbsn} 
\maketitle
\begin{abstract}
Event extraction (EE) is crucial to downstream tasks such as new aggregation and event knowledge graph construction. Most existing EE datasets manually define fixed event types and design specific schema for each of them, failing to cover diverse events emerging from the online text. Moreover, news titles, an important source of event mentions, have not gained enough attention in current EE research. In this paper, We present Title2Event, a large-scale sentence-level dataset benchmarking Open Event Extraction without restricting event types. Title2Event contains more than 42,000 news titles in 34 topics collected from Chinese web pages. To the best of our knowledge, it is currently the largest manually-annotated Chinese dataset for open event extraction. We further conduct experiments on Title2Event with different models and show that the characteristics of titles make it challenging for event extraction, addressing the significance of advanced study on this problem. The dataset and baseline codes are available at \url{https://open-event-hub.github.io/title2event}.
\end{abstract}

\section{Introduction} \label{sec:intro}
\begin{figure}[tbp]
    \centering
    \includegraphics[width=\linewidth]{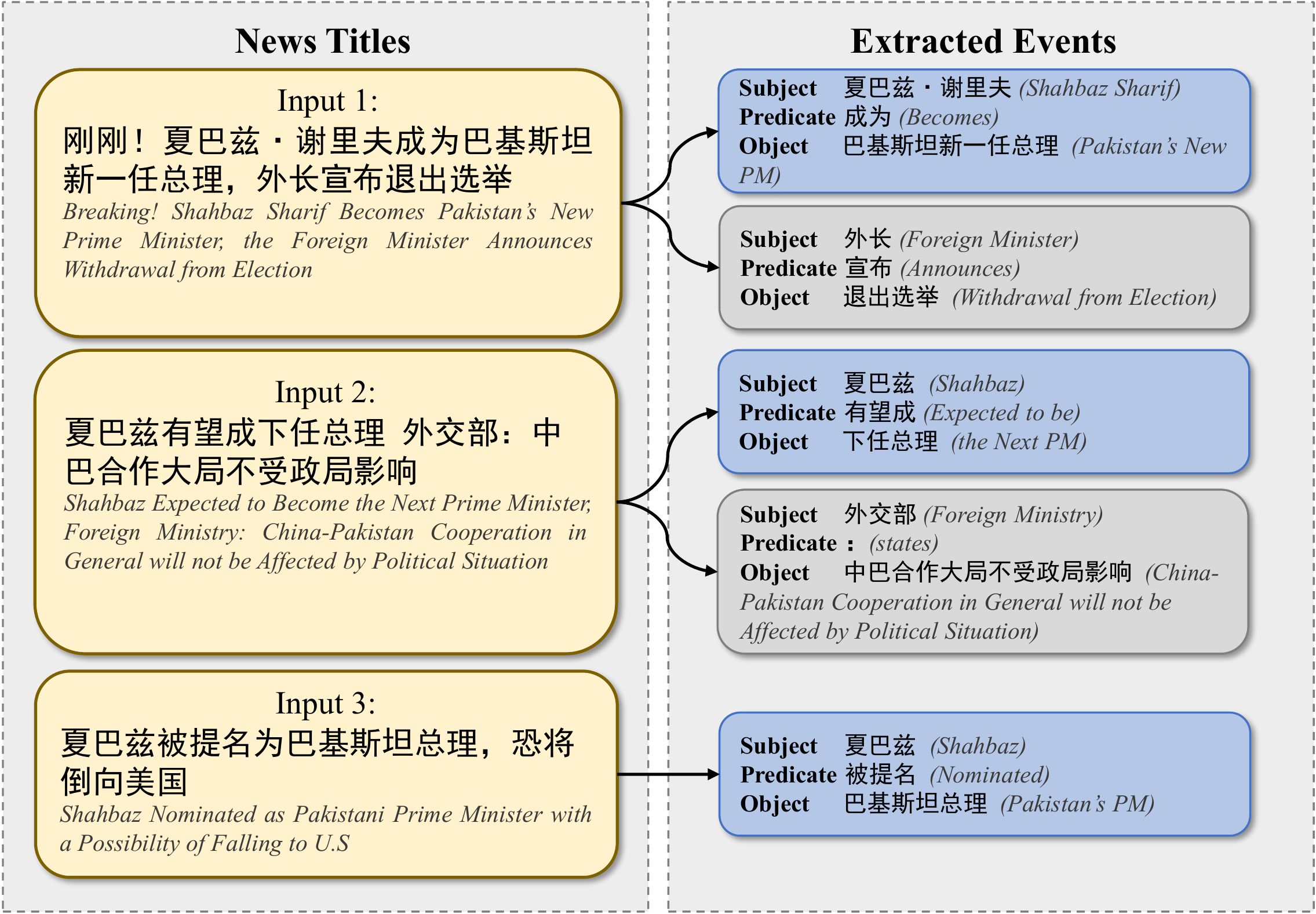}
    \caption{An example of event extraction on news titles where all factual events are extracted. Similar events are identified (in blue color) and could be used in aggregating relevant news.}
    \label{fig:intro}
\end{figure}

Event extraction (EE) is an essential task in information extraction (IE), aiming to extract structured event information from unstructured plain text. Extracting events from news plays an important role in tracking and analyzing social media trending, and facilitates various downstream tasks including information retrieval~\citep{basile2014extending}, news recommendation system~\citep{DBLP:journals/corr/abs-2009-04964} and event knowledge graph construction ~\citep{gottschalk2018eventkg,yu2020enriching,gao2022improving}. Figure~\ref{fig:intro} shows an example of extracting events from multiple news titles. Based on the extracted events, news reporting the same event could be aggregated and sent to users to provide comprehensive views from different sources.

Event extraction can be categorized into two levels: sentence-level EE and document-level EE. Sentence-level EE identifies event entities and attributes in a single sentence~\citep{ahn-2006-stages}, while document-level EE aims to extract entities of the same event scattered across an article~\citep{Sundheim1992OverviewOT}. In scenarios such as news aggregation, human-written news titles often preserve the core information of the news event, while news articles may contain too many trivial details. Therefore, performing sentence-level EE on news titles is more efficient than document-level EE on news articles to aggregate relevant news.

\begin{figure*}[t]
    \centering
    \includegraphics[width=\linewidth]{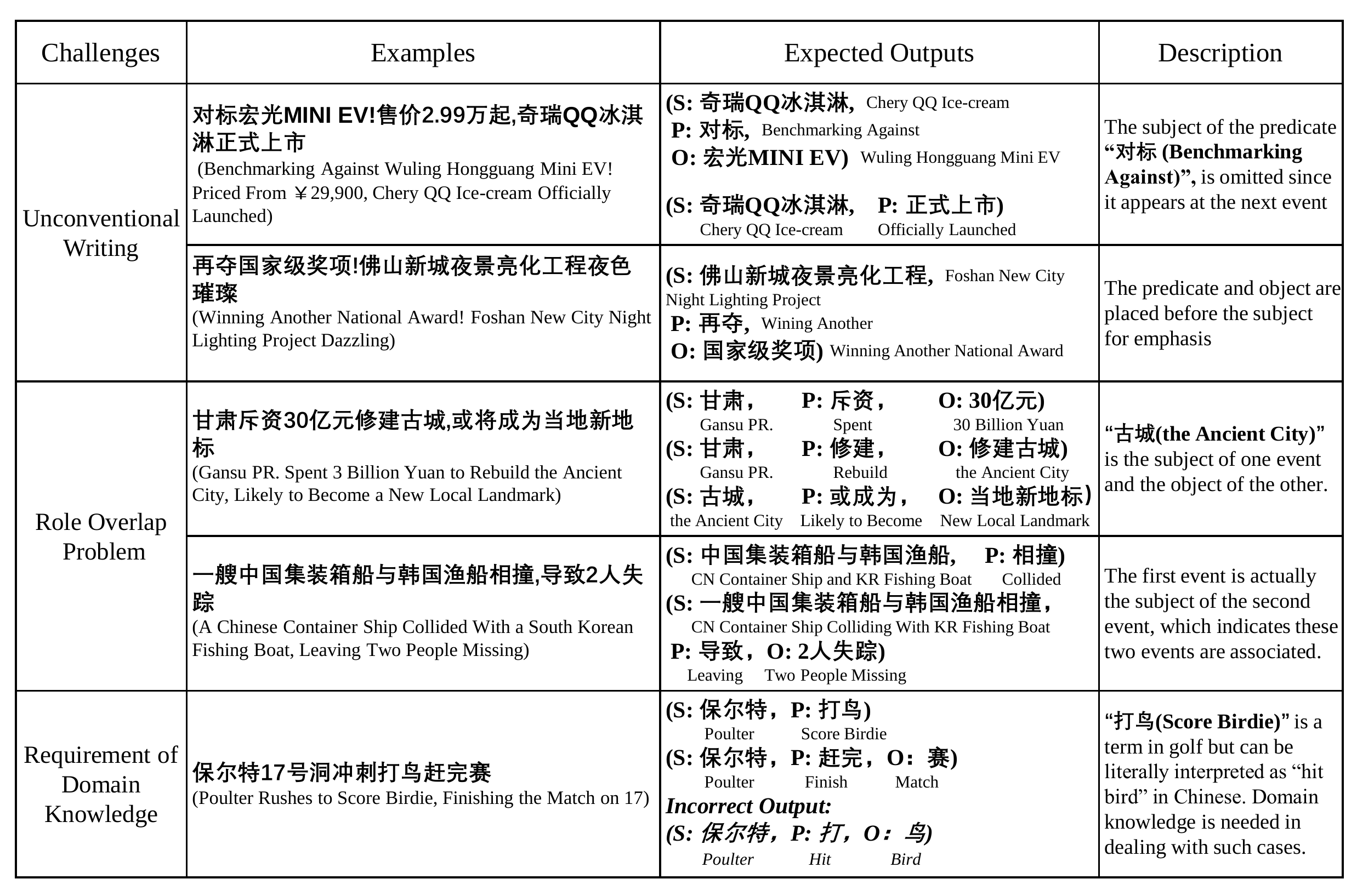}
    \caption{Three types of challenges observed in Title2Event along with their corresponding examples. }
    \label{fig:examples}
\end{figure*}

However, most EE models trained on traditional sentence-level datasets could not reach ideal performance when extracting events from titles \citep{chen-etal-2015-event, Nguyen2019OneFA, Wadden2019ERE, du-cardie-2020-event, li-etal-2020-event, liu-etal-2020-event, lu2021text2event,lou2022translation}. On the one hand, these models request predefined event types and a specific schema for each of them. Each event schema consists of manually designed argument roles such as event trigger, person, time, and location. Then the extraction of events will be decomposed into sub-tasks of extracting each argument role separately. Despite the success in traditional EE, the manual design of specific event schema is costly and time-consuming, and the limited predefined event types could not handle a great variety of events emerging from the Internet where most news titles nowadays are derived from. On the other hand, extracting events from Chinese titles could be more challenging than traditional sentence-level EE such as the ACE 2005 benchmark.\footnote{\url{https://catalog.ldc.upenn.edu/LDC2006T06}} This is because some unique writing styles are observed in news titles on Chinese social media, as shown in Figure~\ref{fig:examples}. First, the writing of many titles does not strictly obey the correct grammar. For example, some titles will omit the agent when describing an action for brevity, while others may place the action before the first mention of the agent for emphasis. Second, the role overlap problem, i.e., the same entity may play different roles in multiple events, usually occurs when the events in the text have certain associations with each other. Although there are about 10\% events in ACE 2005 having this problem, it has not gained enough research attention for quite a long time~\citep{yang2019exploring}. However, the role overlap problem is much more commonly observed in news titles, and thus becomes an issue that can not be ignored. Finally, due to the diverse coverage of news reports, there are some cases in which the EE models have to rely on certain domain knowledge (e.g. rules and terms in sports) for correct event understanding. All these characteristics of titles bring additional challenges to event extraction, demanding EE models of the greater capability of text understanding. 


Considering the significance of title event extraction and a lack of corresponding benchmarks, we present Title2Event, a dataset with more than 42,000 Chinese titles collected from the Internet. In general, Title2Event has the following important features:
\begin{enumerate}
    \item It formulates title event extraction as an open event extraction (OpenEE) task without any predefined event type or specific schema. Instead, it follows the formulation of open information extraction (OpenIE)~\citep{Zhou-2022-OpenIEsurvey} and defines an event as a \texttt{(Subject,Predicate,Object)} triplet. Then, the EE models are required to extract all event triplets in a given title. The biggest difference between OpenEE and OpenIE is that OpenEE is event-centric, which means only triplets of events are to be extracted. 
    \item It is a large-scale, high-quality dataset. Title2Event consists of 42,915 news titles in 34 domains collected from Chinese web pages, along with 70,947 manually annotated event triplets containing 24,231 unique predicates. We write detailed annotation guidelines and conducted two rounds of expert review for quality control. To the best of our knowledge, Title2Event is currently the largest manually annotated Chinese dataset for OpenEE.
    \item It is the first sentence-level dataset with a special focus on titles with its unique values and challenges that little attention has been paid to. We believe Title2Event could further facilitate current EE research in real-world scenarios.  
\end{enumerate}

We experiment with different methods on Title2Event and analyze their performance to address the challenges of this task.

\section{Related Work}
\textbf{Event Extraction Datasets.} 
Automatic Content Extraction (ACE 2005)~\citep{doddington-etal-2004-automatic} is one of the most widely-used corpora in event extraction. It contains 599 documents with 8 event types, 33 event subtypes, and 35 argument roles in English, Arabic and Chinese~\citep{EEsurvey}. 
TAC KBP 2017\footnote{\url{https://tac.nist.gov/2017/KBP/data.html}} is a dataset of the event tracking task in KBP which contains 8 event types and 18 event subtypes in English, Chinese and Spanish. MAVEN~\citep{wang-etal-2020-maven} collects 4,480 Wikipedia documents, 118,732 event mention instances and constructs 168 event types. Despite the large scale, MAVEN merely focuses on event triggers without annotating event arguments. All of the above datasets manually define event types and schema, struggling to handle newly emerging event types in real-world applications.

\noindent \\
\textbf{Open Information Extraction.} 
Open information extraction (OpenIE) aims to extract facts in the form of relational tuples from unstructured text without restricting target relations, relieving human labor of designing complex domain-dependent schema~\citep{niklaus2018survey}. Due to the release of large-scale OpenIE benchmarks such as OIE2016~\citep{stanovsky2016oie} and CaRB~\citep{bhardwaj-etal-2019-carb}, neural OpenIE approaches become popular \citep{Zhou-2022-OpenIEsurvey}. Existing neural OpenIE models can be categorized into sequence tagging models~\citep{stanovsky-etal-2018-supervised, kolluru-etal-2020-openie6, zhan2020span} and generative sequence-to-sequence models~\citep{cui-etal-2018-neural, kolluru2020imojie}. We adopt the formulation of OpenIE and represent events as triplets since the event mentions in news titles tend to be brief without complex substructures. 

\noindent \\
\textbf{Chinese Event Extraction.}
Chinese event extraction can be regarded as a special case of EE due to its unique linguistic properties and challenges~\citep{EEsurvey}.
However, the resources of Chinese EE data are relatively scarce and lack sufficient coverage comparing with EE data in English, which greatly hinders existing research~\citep{Yin-2016-ChEE, lin-etal-2018-nugget, ding-etal-2019-event, Xi-2019-HybridChEE, xu-etal-2020-novel, Cui2020LabelEE}. Apart from multilingual datasets with Chinese corpora such as ACE 2005 and TAC KBP 2017, Chinese Emergency Corpus (CEC)\footnote{\url{https://github.com/shijiebei2009/CEC-Corpus}} collects 6 types of common emergency events. Doc2EDAG~\citep{zheng-etal-2019-doc2edag} and FEED~\citep {li2021feed} are two Chinese financial EE datasets built upon distant supervision. DuEE~\citep{Li-DuEE} is a document-level EE dataset with 19,640 events categorized into 65 event types, collected from news articles on Chinese social media. Compared with DuEE, our Title2Event dataset is larger in scale and does not restrict event types. 

\section{Dataset Construction}
This section describes the process of data collection and annotation details.
\subsection{Data Collection}
We broadly collect Chinese web pages from January to March 2022 using the web crawler logs of the search engine of Tencent as well as a proven business tool to select web pages containing event mentions (most of them are from news websites). Afterwards, the titles of the selected web pages are extracted and automatically tagged with our predefined topics, and titles containing toxic contents are all removed. To ensure the diversity of events, we conduct data sampling every ten days during the crawling period, reducing the occurrence of events belonging to the top frequently appeared topics to make the distribution of topics more balanced. Eventually, around 43,000 instances are collected.

\subsection{Annotation Framework}\label{subsec:annotation}
\noindent
\textbf{Annotation Standard. }
We summarize some essential parts of our annotation standard. In general, we expect each event could be represented by a \texttt{(Subject,Predicate,Object)} triplet where the subject and object could be viewed as the argument roles of the event triggered by the predicate. Multiple event triplets may be extracted from a single title, and they may have some overlaps. However, the predicate of an triplet is considered as the unique identifier of an event, thus multiple triplets of a single title will not share the same predicate. Some important specifications are listed below:

1) We define event as an action or a state of change which occurs in the real world. Some statements such as policy notifications or some subjective opinions are not considered as events. Also, if an title is not clearly expressed, or is concatenated by several unrelated events (e.g. news round-up), then it should be labeled as "invalid" by annotators.

2) We find the identification of predicates in Chinese is complex, so we specify some rules to unify them. First, if an event tends to emphasize the state change of the subject, e.g. “南阳大桥通车”~(Nanyang Bridge opens to traffic), then the predicate will be labeled as “通车”~(open-to-traffic) instead of “通” with “车” as the object. Second, for phrases with serialized verbs and dual objects, we integrate the direct target of the action (i.e. the \textit{Patient}) into the predicate expression while taking the indirect patient (i.e. the \textit{Affectee}) \citep{thompson1973transitivity} as the object of the event. For example, in “送孩子去学校”~(send kids to school) the predicate will be labeled as “送去学校”~(send-to-school) with “孩子”~(kids) as the object. Moreover, we find the colon~("：") frequently plays the role of predicate in titles, representing the meaning of "say", "announce" or "require", etc. We view this as a feature of news titles and allow annotators to label it as the predicate.  
    
3) We expect the fine-grained annotations of argument roles, which are intact yet not redundant. All determiners and modifiers of entities are kept only if they largely affect the understanding of events. All triplets are required to have a subject and a predicate, while the object could be omitted as in the original text.

\noindent \\
\textbf{Crowdsourced Annotation.} 
We cooperate with crowdsourcing companies to hire human annotators. After multi-rounds of training in three weeks, 27 annotators are selected. We pay them ￥1 per instance. Meanwhile, four experts are participated in two rounds of annotation checking for quality control. For each instance, a human annotator is asked to write all expected event triplets independently. To reduce the annotation difficulty, we provide some auxiliary information along with the raw title, including the tokenization outputs, to help annotators quickly capture the entities and concepts present in the titles. Note that we do not force annotators to strictly obey the tokenization outputs, as we find that many of them do not match the desired granularity of triplet elements under our criteria. Instead, the annotation is conducted in a \texttt{<text, label>} pair paradigm rather than a token-level tagging paradigm. Moreover, we provide automatic extraction outputs as references. During the initial phase, we design an unsupervised model to extract triplets. After 20,000 labeled instances are collected, we train a better sequence tagging model for the rest of annotation process. Both models are introduced in Section \ref{sec:methods}. Meanwhile, as titles often contain some domain knowledge which the annotators may not be familiar with, we allow them to refer to search engines. To ensure the quality, we also allow them to label an instance as "not sure" if they are not confident enough. The crowdsourced annotation is conducted in batches. Every batch of annotated instances undergoes two rounds of quality checking before being integrated into the final version of our dataset. We also develop a browser-based web application to accelerate the annotation process, see Appendix~\ref{sec:app_dataset}.

\noindent
\textbf{First-round Checking.}
Each time the crowd-sourced annotation of a batch is completed, it is sent to four experts to check whether they meet the requirements of our annotation standard. Instances which do not pass the quality check will be sent back for revision, attached with the specific reasons for rejection. This process repeats until the acceptance rate reaches 90\%.

\noindent
\textbf{Second-round Checking. }
Each batch of annotated instances passing the first-round checking is sent to the authors for dual check. The authors will randomly check 30\% of the instances and send unqualified instances back to the experts along with the reasons for rejection. Slight adjustments on annotation standard also take place in this phase. This process repeats until the acceptance rate reaches 95\%.

Our annotation process encourages positive interactions among the authors, the experts and the crowdsourced annotators, which effectively helps the annotators to understand the annotation standard and provide timely feedback.

\section{Data Analysis on Title2Event}
This section describes the statistics and characteristics of Title2Event from various perspectives. Table~\ref{tab:overview} shows the overview of the dataset.

\begin{table}[]
    \centering
    \small
    \begin{tabular}[width=\linewidth]{cc} 
    \toprule
    Attribute   &   Count   \\ \midrule
    Data Size (train/val/test)  &  34,295/4,286/4,288 \\ 
    Number of Topics  &  34 \\
    Total Events    &   70,879 \\
    Total Unique Predicates &   24,097 \\
    Avg. Num. of Event per Title & 1.65 \\
    Max. Num. of Event per Title & 6 \\
    Avg. Len. of Title & 23.98 \\
    \bottomrule
    \end{tabular}
    \caption{The overall statistics of Title2Event.}
    \label{tab:overview}
\end{table}

\noindent \\
\textbf{Topic Distribution.}
The titles in the dataset can be categorized into 34 topics, 24 of which contain more than 100 instances. Figure~\ref{fig:topic_dis} lists the distribution of instances belonging to different topics, see Appendix~\ref{sec:app_dataset} for detailed statistics. 

\begin{figure}[htbp]
    \centering
    \includegraphics[width=\linewidth]{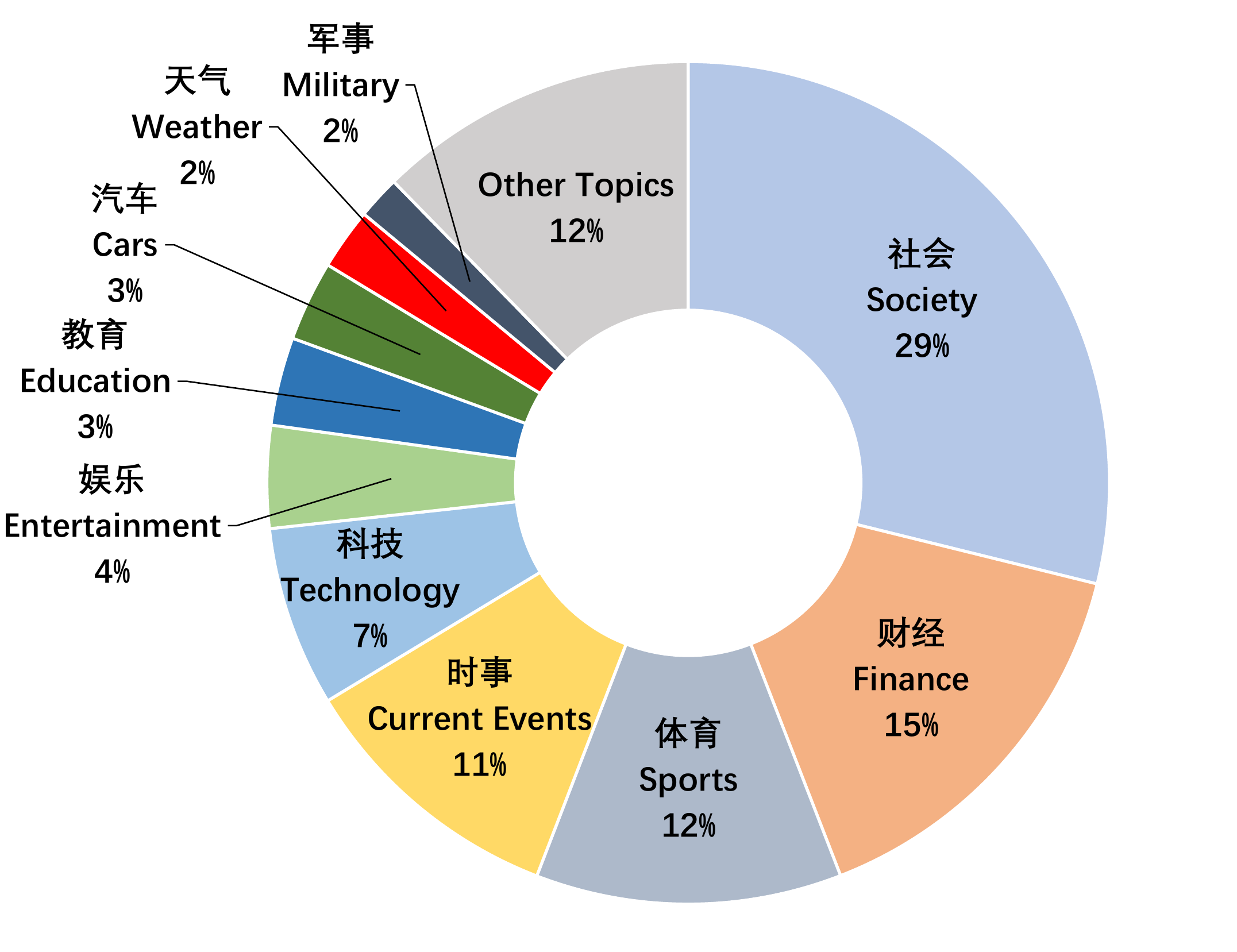}
    \caption{The distribution of topics in Title2Event, all non top-10 topics are aggregated as "Other Topics".}
    \label{fig:topic_dis}
\end{figure}


\begin{figure}
    \centering
    \includegraphics[width=1.0\linewidth]{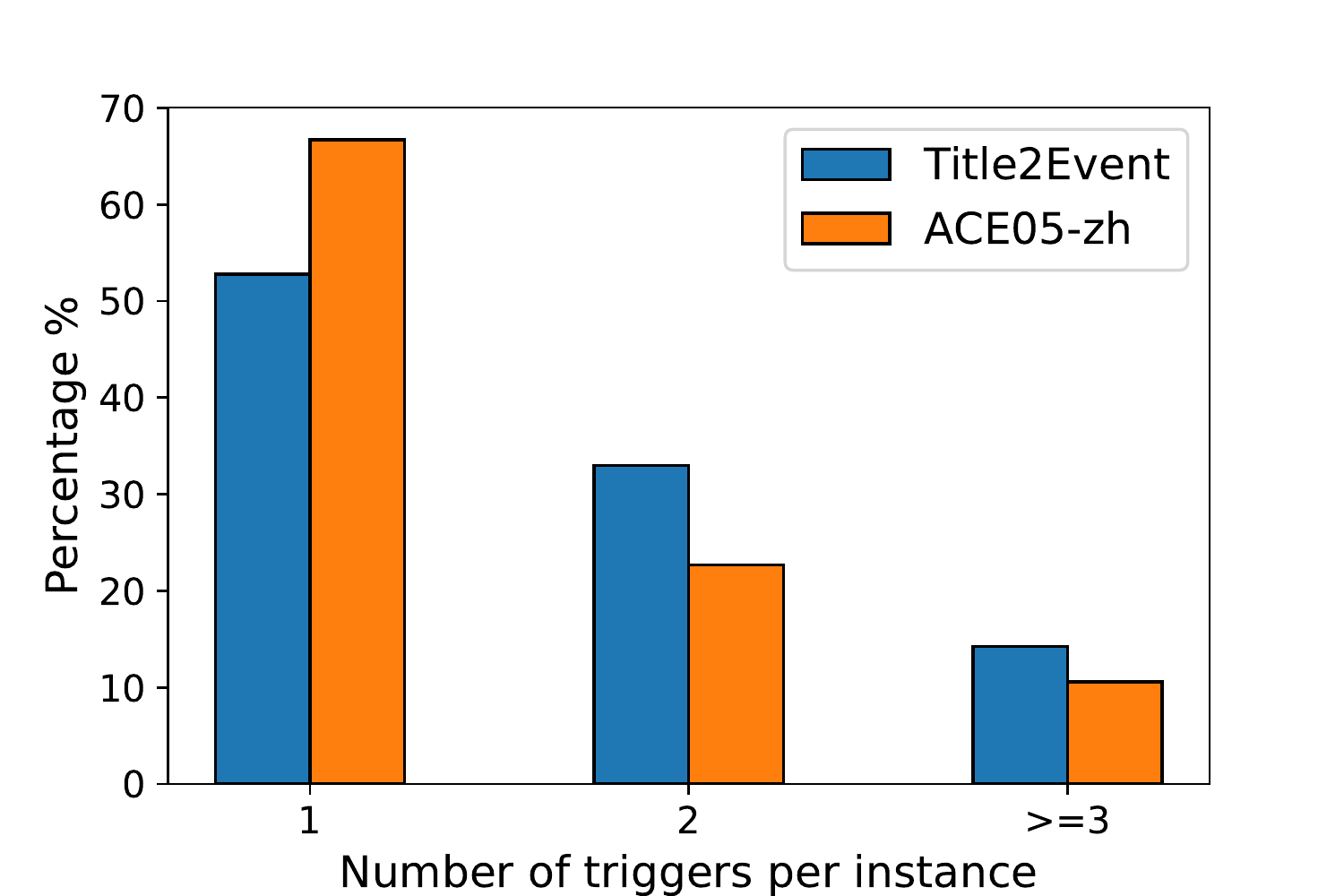}
    \caption{Distribution of instances containing different numbers of triggers of Title2Event and ACE05-zh.}
    \label{fig: trigger_per_ins}
\end{figure}
\begin{figure}
    \centering
    \includegraphics[width=0.9\linewidth]{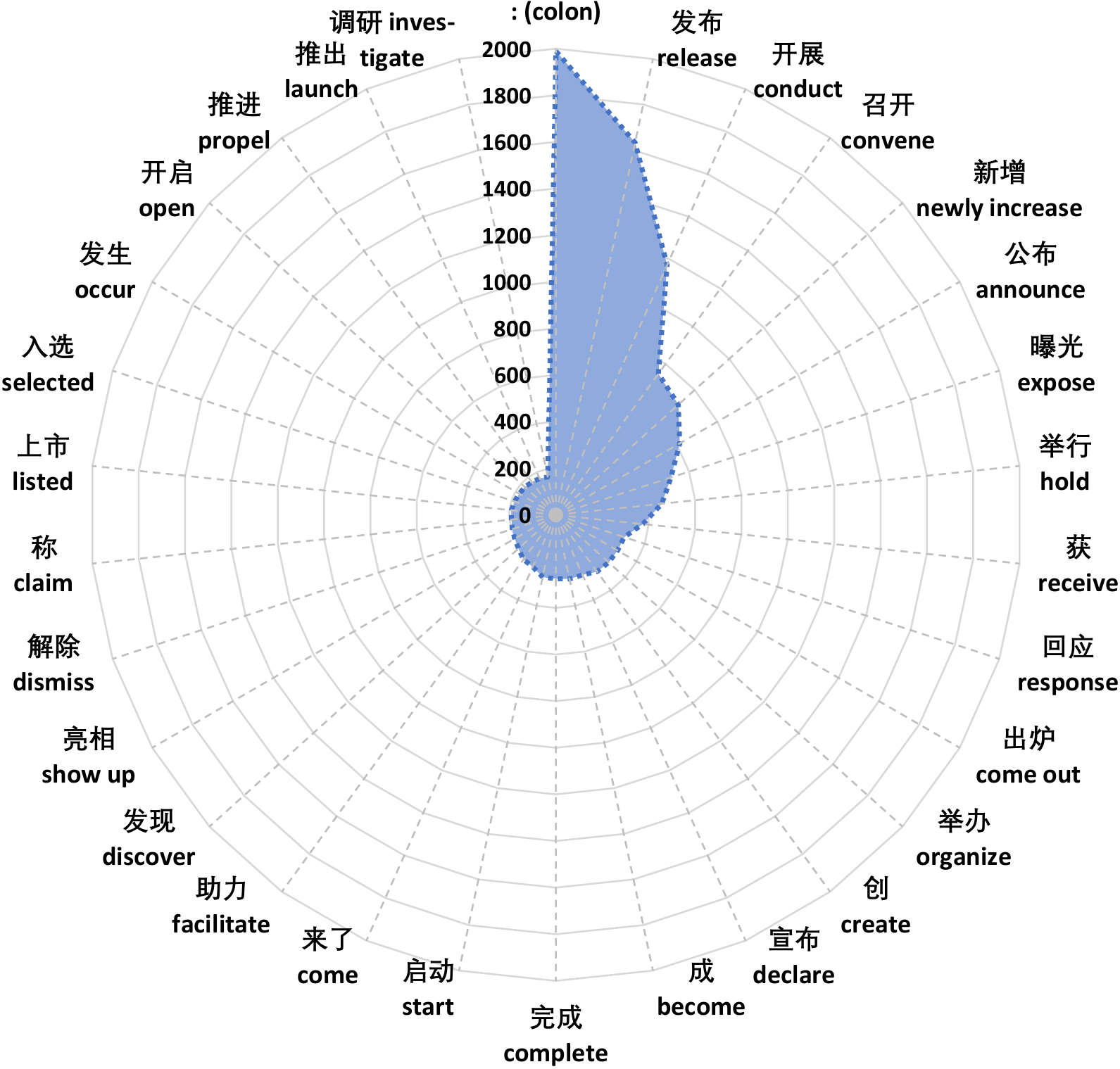}
    \caption{Distribution of top 30 predicates in Title2Event.}
    \label{fig:predicate_dist}
\end{figure}



\noindent \\
\textbf{Event Distribution.}
As shown in Table~\ref{tab:overview}, most of the titles contain more than one event, and the maximum number of events per title is six. We further investigate the distribution of instances containing different numbers of triggers (i.e. predicates for Title2Event), and compare our dataset with the ACE2005 Chinese dataset (denoted as ACE05-zh)
\footnote{We adopt a commonly used preprocessed version of the ACE corpus on sentence level (\url{https://github.com/nlpcl-lab/ace2005-preprocessing}). We also remove all sentences annotated with no event (which accounts for 69.6\% in the processed dataset) for fair comparison.}
as shown in Figure~\ref{fig: trigger_per_ins}. It can be observed that the phenomenon of multiple events per instance is more common in Title2Event compared with ACE05-zh, which brings additional challenges in event extraction.

\noindent \\
\textbf{Predicate Distribution.}
We also investigate the distribution of predicates in Title2Event. Figure~\ref{fig:predicate_dist} shows the distribution of the 30 most frequent predicates in the dataset. 

\noindent \\
\textbf{Challenge Distribution.}
We further analyze to what extent are the observed challenges described in Figure~\ref{fig:examples} covered in Title2Event. To do this, we randomly sample 1,000 instances and manually annotate 1)~Whether the instance omits or inverts some event arguments which makes itself not strictly obeying the grammatical norms. 2)~Whether there's a text span appearing at multiple events of the instance. (3)~Whether some domain knowledge is crucial in understanding the instance that without these knowledge one might not correctly identify the event arguments. The annotation result shows that 9.70\% of sampled instances are observed with unconventional writing, 21.50\% instances have role overlap problem~(10\% for ACE 2005 for comparison), and 2.80\% instances requires domain knowledge for correct event understating. We believe such statistics are a good identification of the challenging nature of Title2Event.
        
\begin{table*}[htbp]
\centering
\begin{tabular}{cccccccccc}
\toprule
Methods    & \multicolumn{3}{c}{Trigger Ex.} & \multicolumn{3}{c}{Argument Ex.} & \multicolumn{3}{c}{Triplet Ex.} \\
            & P      & R      & F1     & P      & R      & F1     & P      & R      & F1     \\
\midrule
Unsuper.        & 21.0   & 32.0   & 25.4   & 12.0   & 15.5   & 13.5   & 4.5    & 6.8    & 5.4       \\
\hline
SeqTag          & \textbf{69.5}   & \textbf{69.9}   &\textbf{69.7}   & 50.8   & 51.2   & 51.0   & 41.1   & 41.3   & 41.2  \\
ST-SpanMRC   & -      & -      & -      & \textbf{60.1}   & 54.9   & 57.4   & 44.5   & 44.8   & 44.7   \\
ST-Seq2SeqMRC  & -    & -     & -      & 57.9   & \textbf{58.6}    & \textbf{58.2}   & \textbf{49.8}  & \textbf{50.1}    & \textbf{49.9}   \\         
\bottomrule
\end{tabular}
\caption{The overall results of trigger extraction~(Trigger Ex.), argument extraction~(Argument Ex.), and event triplet extraction~(Triplet Ex.) on Title2Event. P, R, F1 stand for precision, recall, and f1-score respectively.}
\label{tab:main_results}
\end{table*}

\section{Methods}\label{sec:methods}
Formally, given a sequence of tokens $S=\ <w_1,w_2,\ldots, w_n>$, Open EE aims to output a list of triplets $T=<t_1, t_2, \ldots, t_m>$ where each triplet $t_i=<s_i, p_i, o_i>$ represents an event occurred in $S$ and $s_i, p_i, o_i$ denote the subject, predicate and object of the event respectively. The object of an event could be empty, and the total number of events per sentence $m$ is not fixed. Open EE can also be aligned with traditional EE task formulation by considering the predicate as the event trigger as well as a unique event type, while the subjects and objects both taken as event arguments.

Based on the task formulation, we first implement an unsupervised method using an existing toolkit. Then, we split the task into trigger extraction and argument extraction, and implement different supervised methods on them.

\subsection{Unsupervised Method}
Since the formulation of Open EE is similar to some traditional tasks such as dependency parsing~(DP) and semantic role labeling~(SRL), we investigate the performance of existing triplet extraction methods on Open EE. Each title will be segmented and tokenized first, then the extraction is conducted as a token-wise sequence-labeling task. Each token will first be labeled by a SRL module on whether it belongs to a semantic role which appears in one of the S-P-O, S-P, P-O semantic tuples. If not, it will be relabeled by a DP module on whether it appears in a syntactical tuple of the above structures. The entire method is implemented using the LTP toolkit~\citep{che2020n}. 

\subsection{Trigger Extraction}
Since the number of triggers per sentence is neither fixed nor given as input, we adopt a token-level sequence tagging model to extract all event triggers in a given sentence based on the inductive bias that event triggers (i.e., predicates) will not overlap with each other (see Section~\ref{subsec:annotation}). Sequence tagging model requires a set of tags where each tag, aligned with a token, represents a part of an event element (i.e., a triplet element) or a non-event element. Then, the model learns the probability distributions of tags for each given sentence, and outputs triplets based on the predicted tags. We adopt the BIO tagging scheme where a token is tagged \texttt{$B\text{-}trg_i$} (\texttt{$I\text{-}trg_i$}) if it is at the beginning of (inside) the \textit{$i^{th}$} trigger, or \texttt{$O$} if it is outside any trigger. The subscript is used to distinguish between different triggers as they might be discontinuous tokens. Since Title2Event is not annotated on token-level (see \ref{subsec:annotation}), we perform automatic tagging by locating each annotated event element at the source sentence to get its offset. We use BERT~\citep{devlin2018bert} as the sentence encoder to get the contextualized representations of tokens, and each token representation will be fed to a classification layer to compute the probability distribution of the tags. 



\subsection{Argument Extraction}
Argument extraction models take the source sentence and the given triggers as input and output the arguments of each given trigger respectively. Due to the role overlap problem, a token might appear in multiple event arguments and thus has multiple tags, which does not match the common setting of sequence tagging task. Therefore, we iterate over the extracted triggers and extract the arguments of each event trigger separately. We implement three methods for argument extraction. 

\noindent \\
\textbf{Sequence Tagging. }
The first method is a \textbf{token-level sequence tagging model} similar to the trigger extraction model, which also uses BIO tagging scheme for subject and object tokens. During each forward process, to specify the current trigger, we adopt the method proposed by \newcite{yang2019exploring}. Specifically, the input of BERT encoder is the sum of WordPiece embeddings, position embeddings and segment embeddings, and we set the segment ids of current trigger tokens being one while others being zero to explicitly encode the current trigger. 

\noindent 
\textbf{Span MRC. }
The second method is a \textbf{span-level tagging model} which formulates argument extraction as a machine reading comprehension (MRC) task, inspired from \newcite{du-cardie-2020-event} and \newcite{liu-etal-2020-event}. For each given sentence as well as a specified trigger, the subject and object are extracted separately by prepending a question, e.g. “动作\texttt{<trigger>}的主体是？”~(What is the subject of \texttt{<trigger>)}?, into the sentence to form a context like \texttt{"[CLS] question [SEP] sentence [SEP]"}, then the model is asked to extract the answer span from the context for the given question by predicting a start position and an end position. We also use BERT as the context encoder.   

\noindent \\
\textbf{Seq2Seq MRC. }
The third method is a sequence-to-sequence MRC model with same the question design as \textbf{Span MRC}. However, instead of extracting the answer spans from the context, it directly generates a sequence of tokens as the output with the given context by maximizing the conditional probability $P(Y \mid S)=\prod_{i=1}^{m} p\left(y_{i} \mid y_{1}, y_{2}, \ldots, y_{i-1} ; S\right)$, where $Y=<y_1, \ldots, y_m>$ is the golden answer. We adopt mT5~\citep{xue2020mt5}, a multilingual text-to-text transformer model as the context encoder as well as the answer decoder.



\section{Experiments} \label{sec:exp}
We conduct experiments on Title2Event with the methods described in Section~\ref{sec:methods} and analyze their performance.

\begin{table}[]
    \centering
    \small
    \setlength\tabcolsep{4pt}
    \begin{tabular}{ccccccc}
\toprule
Methods & \multicolumn{3}{c}{Argument Ex.} & \multicolumn{3}{c}{Argument Ex.~(Gold)}   \\
        & P     & R     & F1        & P     & R     & F1  \\ 
\midrule
SeqTag  & 50.8   & 51.2   & 51.0    & 70.4  & 69.6  & 70.0  \\
SpanMRC  & 60.1   & 54.9   & 57.4    & 82.9  & 74.8  & 78.6  \\
Seq2SeqMRC & 57.9 & 58.6   & 58.2    & 80.6  & 80.4  & 80.5  \\
\bottomrule
    \end{tabular}
    \caption{Results of argument extraction with predicted triggers~(Argument Ex.) and with golden triggers~(Argument Ex.~(Gold))}
    \label{tab:pipeline-error}
\end{table}

\subsection{Evaluation Metrics}
We adapt the evaluation metrics used in previous works on traditional EE tasks~\citep{EEsurvey} to Open EE. 
We first define the matching criteria: an event trigger or argument is correctly identified if it exactly matches the golden answer, and an event triplet is correctly identified only if all of its elements are correctly identified. We then compute the precision~(P), recall~(R), and F1-score~(F1) for trigger extraction, argument extraction and triplet extraction respectively.

\subsection{Evaluation Model}
We summarize all the models we implement for experiments here:

\noindent  \textbf{Unsuper. } The unsupervised triplet extraction method implemented by the LTP toolkit using the Chinese-ELECTRA-small~\citep{cui-etal-2021-pretrain} model.

\noindent \textbf{SeqTag. } A pipeline tagging-based model consisting of a trigger extractor and an argument extractor, both are based on the token-level sequence tagging model using BERT-base-Chinese as the encoder, and are trained separately. During inference, the argument extractor predicts the arguments based on the triggers predicted by the trigger extractor.

\noindent \textbf{ST-SpanMRC. } A pipeline model using a token-level sequence tagging model as the trigger extractor, and a span-level MRC model as the argument extractor, both are based on BERT-base-Chinese.

\noindent \textbf{ST-Seq2SeqMRC. } A pipeline model which replaces the argument extractor with a sequence-to-sequence MRC model using mT5-base.

\begin{figure*}[htbp]
    \centering
    \includegraphics[width=\linewidth]{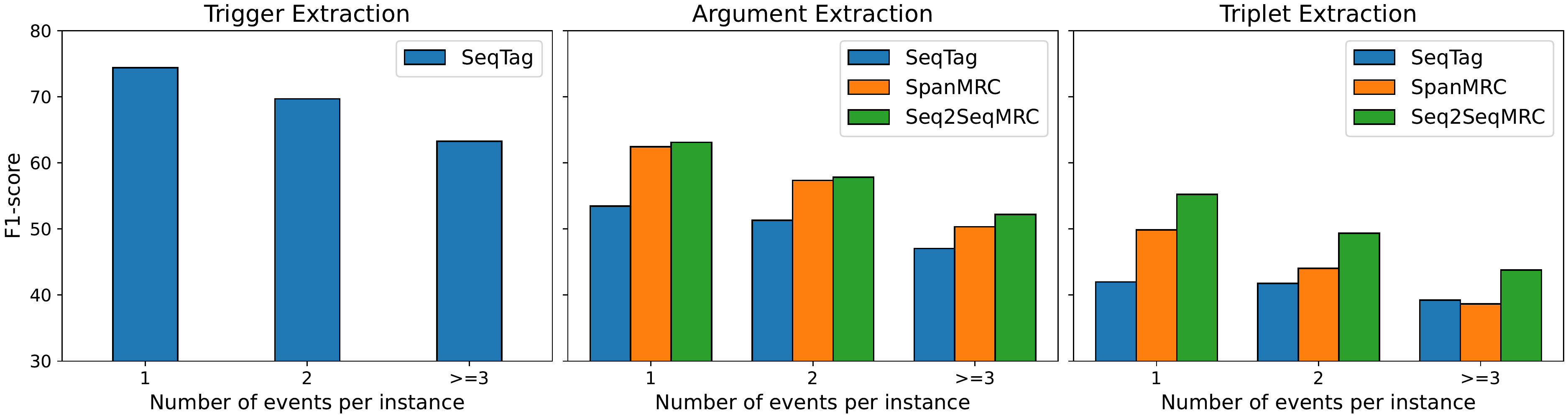}
    \caption{Results of event extraction on instances containing different number of events.}
    \label{fig:result_analysis}
\end{figure*}

\subsection{Overall Experimental Results}
Table~\ref{tab:main_results} shows the results of all Open EE methods experimented on Title2Event. It can be observed that: 1) For trigger extraction, the sequence tagging model significantly outperforms the unsupervised model. 2) For argument extraction and triplet extraction, ST-Seq2SeqMRC outperforms the other tagging-based models. A large part of the reason is that the unconventional writing styles of titles make it difficult to locate token-level tags or span offsets in the source text, while sequence-to-sequence models are free from these restrictions. 

\subsection{Analysis on Error Propagation} \label{subsec:pipeline_error}
Table~\ref{tab:pipeline-error} shows the results of argument extraction with predicted triggers and with golden triggers. All three models' performance improve by approximately 20\% if provided with golden triggers, indicating the huge impact of correct triggers on argument extraction and the urgent need to alleviate the propagating error brought by pipeline architecture in future works.   

\subsection{Analysis on Multiple Event Extraction}
Figure~\ref{fig: trigger_per_ins} shows that containing multiple events per instance is an important feature of Title2Event, thus we further investigate the models' performance on multiple event extraction, as shown in Figure~\ref{fig:result_analysis}. We can see that as the number of events per instance increases, all models on trigger extraction, argument extraction, and triplet extraction 
show a decrease in performance, which indicates that multiple events per instance brings additional challenges to open event extraction.

\begin{figure}[p]
    \centering
    \includegraphics[width=\linewidth]{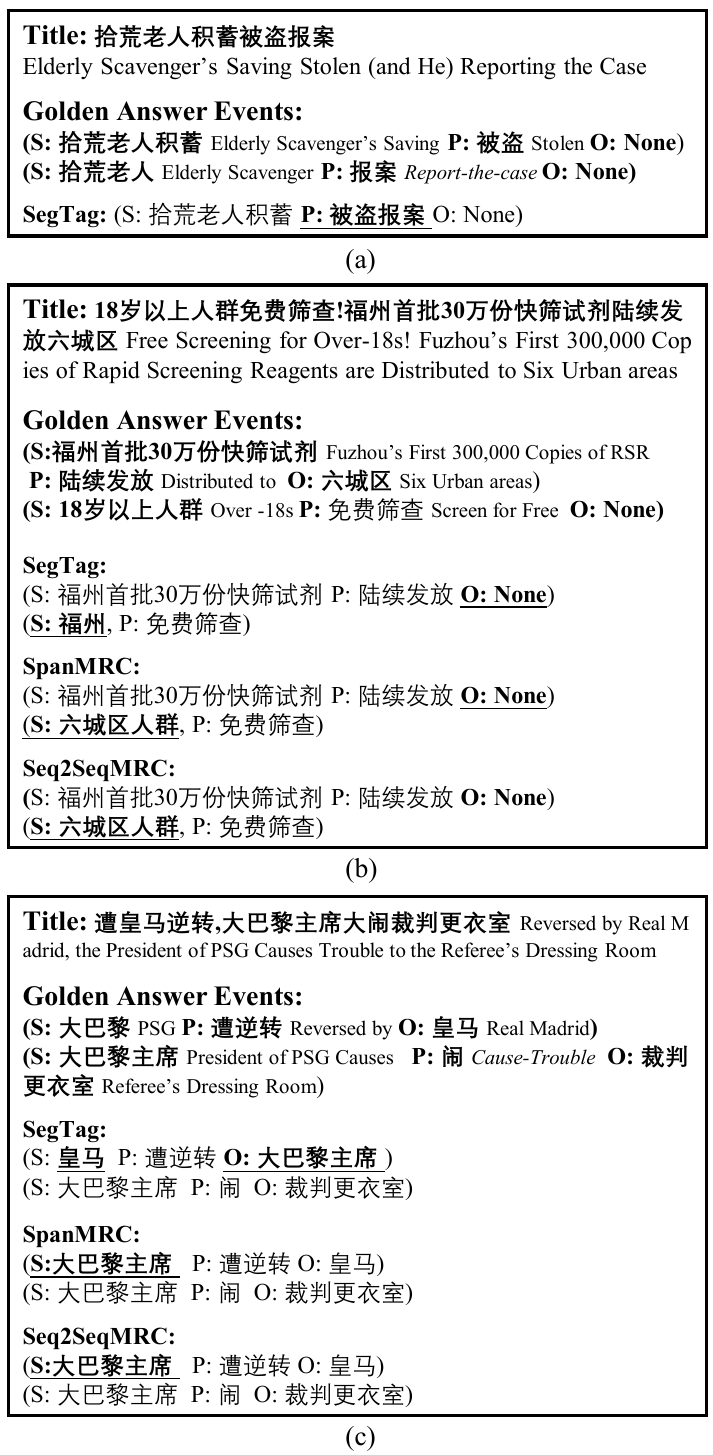}
    \caption{Example error cases, arguments in bold text and underlined are the specific errors compared to golden answers.}
    \label{fig:cases}
\end{figure}


\subsection{Analysis on Different Topics}
We also investigate the results of trigger extraction and argument extraction on different topics of Title2Event, see Appendix~\ref{sec:app_dataset} for details. It can be observed that the F1-scores of "Weather" are higher than other topics, probably because news titles on weather (forecast) usually have a fixed template which makes extraction easier.

\subsection{Analysis on Error Cases}
We summarize three typical challenges observed in Title2Event in Section~\ref{sec:intro}. Here, we analyze some error cases of the model outputs to further demonstrate the issues. Figure~\ref{fig:cases}~(a) shows an error output in trigger extraction, where the given title is unconventionally written by concatenating two predicates. As a result,
SegTag is unable to distinguish the two different predicates. Figure~\ref{fig:cases}~(b) shows an instance with multiple events and all the models mix up the argument roles. Figure~\ref{fig:cases}~(c) shows a sport news title, without the background that \textit{Real Madrid} and \textit{PSG} are both football clubs, none of the models properly understand the event that \textit{PSG} is defeated by \textit{Real Madrid}. All of the above cases clearly address the challenges present in Title2Event, which are also common in real-world scenarios, and require advanced study to be better solved.

\section{Conclusions}
In this paper, we present Title2Event, a Chinese title dataset benchmarking the task of open event extraction. To the best of our knowledge, Title2Event is the largest manually-annotated Chinese dataset for sentence-level event extraction. We experiment with different methods and conduct detailed analysis to address the challenges observed in Title2Event, which are rather scarce in existing datasets yet common in real-world scenarios. We believe Title2Event could further facilitate advanced research in event extraction.  

\section*{Limitations}
We summarize the limitations of Title2Event as follows: \\
\textbf{Evaluation Metrics.} We make Title2Event a benchmark for open event extraction with a hope that it could evaluate the performance of domain-general EE models. We adapt the formulation of Open IE and represent events in a universal triplet format while adopting traditional EE metrics which is based on exact match. However, we observe that the narrative of events in Chinese titles are extremely diverse. To unify them into the triplet format without losing the core event information, we design detailed annotation guidelines which results in the fact that the a large amount of triplet elements are text spans instead of one or two tokens which is common in traditional EE datasets such as ACE 2005. Therefore, using exact match in Title2Event might be too strict for model outputs which are just one or two tokens different from the golden text span. We leave the design of fine-grained evaluation approaches to future work.  

\noindent \\
\textbf{Methods.} Some characteristics of Title2Event such as unfixed number of events per instance and the role overlap problem bring difficulties to the model design. We adopt a pipeline architecture which suffers from the error propagation problem as discussed in Section~\ref{subsec:pipeline_error}. We also adapt some end-to-end models in traditional EE such as TEXT2EVENT proposed by \newcite{lu2021text2event} to our Open EE benchmark, but find the performance is unexpectedly poor. We conduct preliminary analysis and find that the length of text span in triplets~(as mentioned above) as well as the relatively complex linearized event structures~(largely due to the multiple events per instance issue) are the potential factors of the limited performance. Therefore, we do not provide a good end-to-end model as baseline, which might make the model comparison in Section \ref{sec:exp} less comprehensive. However, we hope that future works could pay more attention to the design of text-to-structure models except from traditional tagging-based models. 



\section*{Ethics Statement}
As Title2Event is an Open EE dataset which broadly collects contents of various categories on the Internet, keeping the corpus without bias is extremely difficult. However, we put large efforts in cleaning the toxicity of data. First, all crawled web paged are automatically removed if they contain toxic contents using an existing system. During annotation, all instances will be dual checked by the human annotators and manually deleted if not passing the check. Moreover, in our annotation standard, we ask annotators to label only factual events while ignoring all subjective opinions, as we hope Title2Event could be factual and unbiased. 



\bibliography{anthology, myref}
\bibliographystyle{acl_natbib}

\clearpage

\appendix
\section{Appendix}\label{sec:app_dataset}

\begin{figure*}[t]
    \centering
    \includegraphics[width=\linewidth]{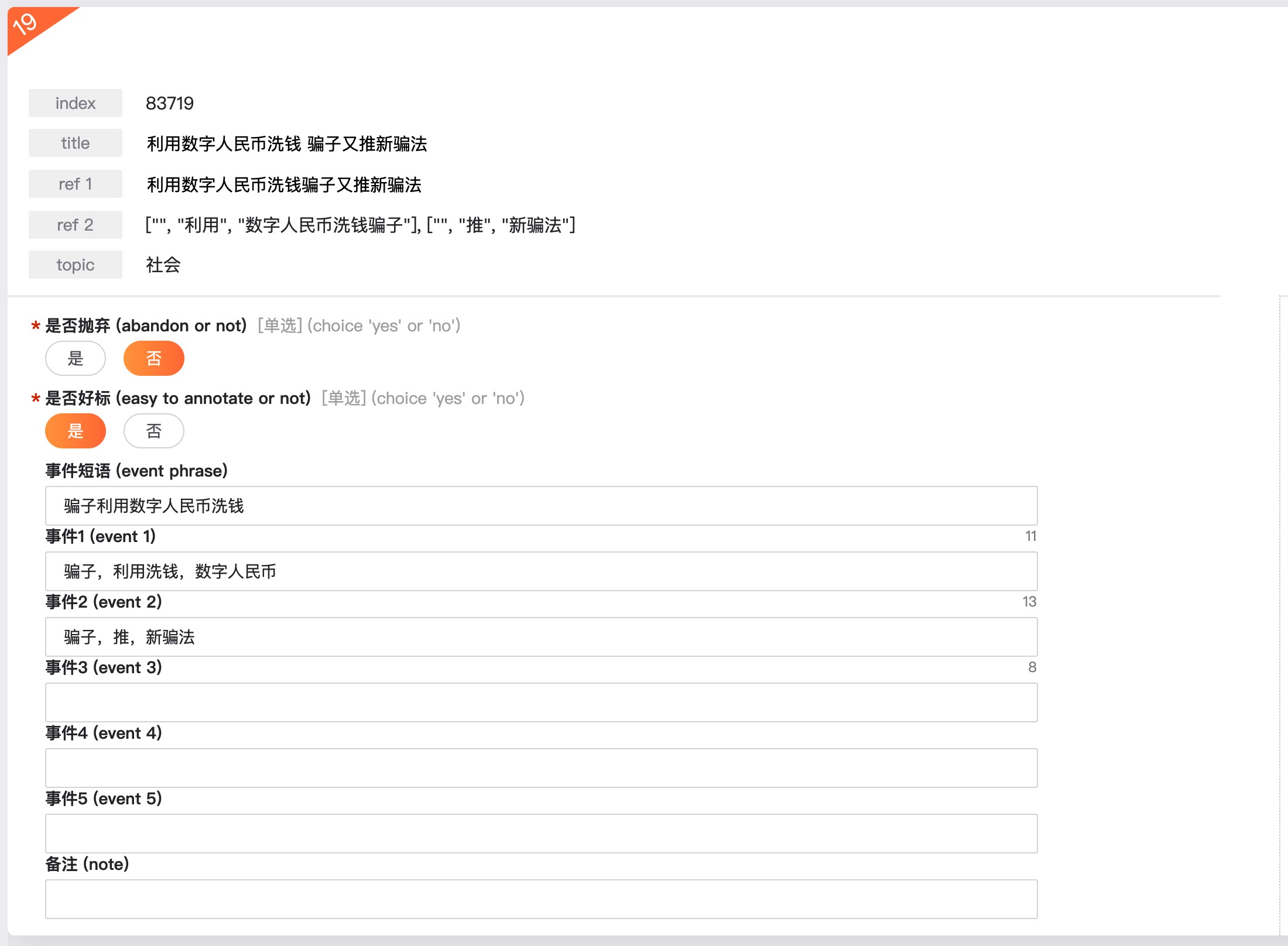}
    \caption{Screenshot of our annotation web page.}
    \label{fig:annotation_tool}
\end{figure*}

\textbf{Annotation Tool.}
Figure~\ref{fig:annotation_tool} shows a screenshot of our annotation web page. The raw title are given with auxiliary information, the annotators will first determine whether to abandon this case as well as is this case easy to annotate or not. Then, they will type all plausible events in the text boxes following our annotation guidelines.

\noindent \\
\textbf{Topic List.}
Table~\ref{tab:topic_dis} lists all 34 topics and their corresponding number of instances.

\begin{table}[h]
    \centering
    \begin{tabular}{lc} 
    \toprule
    Topic & Count \\
    \midrule
    社会~(Society) & 12,341 \\
    财经~(Finance) & 6,539 \\
    体育~(Sports)   & 5,033 \\
    时事~(Current Events) & 4,499 \\
    科技~(Technology) & 2,965 \\
    娱乐~(Entertainment) & 1,685 \\
    教育~(Education) & 1,451 \\
    汽车~(Cars) & 1,319 \\
    天气~(Weather) & 1,013 \\
    军事~(Military) & 712 \\
    旅游~(Travel) & 659 \\
    房产~(Real Estate) & 647 \\
    三农~(Agriculture) & 520 \\
    文化~(Culture) & 501 \\
    综艺~(Variety Shows) & 412 \\
    游戏~(Games) & 396 \\
    电影~(Movies) & 348 \\
    健康~(Health) & 344 \\
    电视剧~(TV Series) & 233 \\
    历史~(History) & 220 \\
    音乐~(Music) & 159 \\
    科学~(Science) & 147 \\
    生活~(Life) & 118 \\
    美食~(Food) & 117 \\
    情感~(Sentiment) & 95 \\
    育儿~(Childcare) & 73 \\
    时尚~(Fashion) & 60 \\
    宠物~(Pets) & 57 \\
    职场~(Career) & 54 \\
    曲艺~(Folk Art) & 41 \\
    动漫~(Animation) & 34 \\
    摄影~(Photography) & 24 \\
    搞笑~(Funny News) & 12 \\
    其它~(Others) & 11 \\
    \bottomrule
    \end{tabular}
    \caption{The topics in Title2Event with their number of instances.}
    \label{tab:topic_dis}
\end{table}

\begin{figure*}
    \centering
    \includegraphics[width=\linewidth]{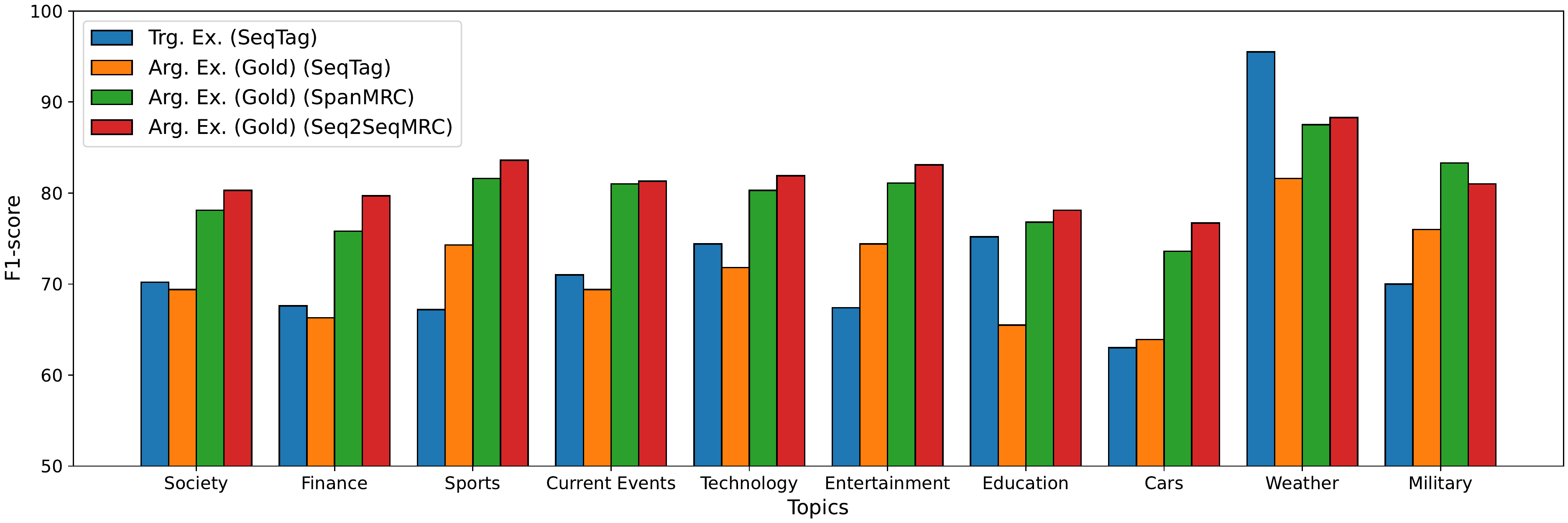}
    \caption{Results of trigger extraction and argument extraction with golden triggers on top 10 topics in Title2Event}
    \label{fig:topic_result}
\end{figure*}

\noindent \\
\textbf{Results on Different Topics.}
Figure~\ref{fig:topic_result} shows the F1-scores of trigger extraction (using SeqTag model) and argument extraction with golden triggers (using SeqTag, SpanMRC, and Seq2SeqMRC models) on the top-10 topics in Title2Event.

\noindent \\
\textbf{Hyper-parameter Settings in Training.}
For all models, we use the batch size of 32 and train them for 30 epochs on the training set of Title2Event. All models are trained on a single Tesla A100 GPU. We use the linear learning rate scheduler and AdamW as the optimizer. For models based on Bert-base-Chinese, we set the learning rate to be 5e-5; For models based on mT5-base, we set the learning rate to be 1e-4. All supervised models are implemented using the Huggingface-transformers library.

\end{CJK*}
\end{document}